\documentclass[11pt,a4paper]{article}
\usepackage[hyperref]{acl2021}
\usepackage{times}
\usepackage{latexsym}

\usepackage{microtype}
\usepackage{tabularx}
\usepackage{booktabs}
\usepackage{multirow}
\usepackage{color}
\usepackage{amssymb}
\usepackage{bm}
\usepackage{graphicx}
\usepackage{amsmath}

\aclfinalcopy 

\title{GTM: A Generative Triple-Wise Model for \\ Conversational Question Generation}

\def\first{$^1$}
\def\second{$^2$}
\def\third{$^3$}
\def\comma{$^,$}
\def\star{$^*$}

\author{Lei Shen\first\comma\second 
~~~ Fandong Meng\third
~~~ Jinchao Zhang\third
~~~ Yang Feng\first\comma\second\star
~~~ Jie Zhou\third
\\
{\first {Institute of Computing Technology, Chinese Academy of Sciences, Beijing, China}}  \\
{\second {University of Chinese Academy of Sciences, Beijing, China}} \\
{\third {Pattern Recognition Center, WeChat AI, Tencent Inc, China}} \\
{\tt \small{shenlei17z@ict.ac.cn}}, {\tt \small{ \{fandongmeng,dayerzhang\}@tencent.com}}\\ 
{\tt \small{fengyang@ict.ac.cn}}, {\tt \small{withtomzhou@tencent.com}}}

\date{}

\begin{document}
\maketitle
\newcommand\blfootnote[1]{%
\begingroup 
\renewcommand\thefootnote{}\footnote{#1}%
\addtocounter{footnote}{-1}%
\endgroup
}
\blfootnote{Joint work with Pattern Recognition Center, WeChat AI, Tencent Inc, China. \star{Yang Feng is the corresponding author.}}

\begin{abstract}
Generating some appealing questions in open-domain conversations is an effective way to improve human-machine interactions and lead the topic to a broader or deeper direction. To avoid dull or deviated questions, some researchers tried to utilize answer, the ``future'' information, to guide question generation. However, they separate a post-question-answer (PQA) triple into two parts: post-question (PQ) and question-answer (QA) pairs, which may hurt the overall coherence. Besides, the QA relationship is modeled as a one-to-one mapping that is not reasonable in open-domain conversations. To tackle these problems, we propose a generative triple-wise model with hierarchical variations for open-domain conversational question generation (CQG). Latent variables in three hierarchies are used to represent the shared background of a triple and one-to-many semantic mappings in both PQ and QA pairs. Experimental results on a large-scale CQG dataset show that our method significantly improves the quality of questions in terms of fluency, coherence and diversity over competitive baselines.
\end{abstract}

\section{Introduction}
\label{sec:intro}

\begin{table}[htb]
\small
\centering
\begin{tabular}{p{5.5cm}}
  \toprule[1pt]
  \textbf{Post:} \\
  I ate out with my friends this evening. \\
  \midrule[1pt]
  \textbf{Question Candidates: } \\
  Q1.1: \textcolor[rgb]{0,0,0.8}{Which restaurant} did you go? \\
  Q1.2: \textcolor[rgb]{0,0,0.8}{Where} did you eat? \\
  Q2.1: What \textcolor[rgb]{0,0.5,0}{food} did you \textcolor[rgb]{0,0.5,0}{eat}? \\
  Q2.2: Did you \textcolor[rgb]{0,0.5,0}{eat something} special? \\
  Q3: What do you mean? \\
  Q4: How about drinking together? \\
  \midrule[1pt]
  \textbf{Answer Candidates: } \\
  A1: We went to an \textcolor[rgb]{0,0,0.8}{Insta-famous cafeteria}. \\
  A2: We \textcolor[rgb]{0,0.5,0}{ate steak and pasta}. \\
  \bottomrule[1pt]
\end{tabular}
\caption{An example of CQG task which is talking about a person's eating activity. There are one-to-many mappings in both PQ and QA pairs. The content of each meaningful and relevant question (Q1.1 to Q2.2) is decided by its post and answer. Q3 (dull) and Q4 (deviated) are generated given only the post.}
\label{tab:intro}
\end{table}

Questioning in open-domain dialogue systems is indispensable since a good system should have the ability to well interact with users by not only responding but also asking \cite{li2016learning}. Besides, raising questions is a proactive way to guide users to go deeper and further into conversations \cite{yu2016strategy}. Therefore, the ultimate goal of open-domain conversational question generation (CQG) is to enhance the interactiveness and maintain the continuity of a conversation \cite{wang2018learning}. CQG differs fundamentally from traditional question generation (TQG) \cite{zhou2019question,kim2019improving,li2019improving} that generates a question given a sentence/paragraph/passage and a specified answer within it. While in CQG, an answer always follows the to-be-generated question, and is unavailable during inference \cite{wang2019answer}. At the same time,  each utterance in open-domain scenario is casual and can be followed by several appropriate sentences, i.e., one-to-many mapping \cite{gao2019jointly,chen2019generating}.

At first, the input information of CQG was mainly a given post \cite{wang2018learning,hu2018aspect}, and the generated questions were usually dull or deviated (Q3 and Q4 in Table \ref{tab:intro}). Based on the observation that an answer has strong relevance to its question and post, \citet{wang2019answer} tried to integrate answer into the question generation process. They applied a reinforcement learning framework that firstly generated a question given the post, and then used a pre-trained matching model to estimate the relevance score (reward) between answer and generated question. This method separates a post-question-answer (PQA) triple into post-question (PQ) and question-answer (QA) pairs rather than considering the triple as a whole and modeling the overall coherence. Furthermore, the training process of the matching model only utilizes one-to-one relation of each QA pair and neglects the one-to-many mapping feature.

An open-domain PQA often takes place under a background that can be inferred from all utterances in the triple and help enhance the overall coherence. When it comes to the semantic relationship in each triple, the content of a specific question is under the control of its post and answer \cite{lee2020generating}. Meanwhile, either a post or an answer could correspond to several meaningful questions. As shown in Table \ref{tab:intro}, the triple is about a person's eating activity (the background of the entire conversation). There are one-to-many mappings in both PQ and QA pairs that construct different meaningful combinations, such as P-Q1.1-A1, P-Q1.2-A1, P-Q2.1-A2 and P-Q2.2-A2. An answer connects tightly to both its post and question, and in turn helps decide the expression of a question.

On these grounds, we propose a generative triple-wise model (GTM) for CQG. Specifically, we firstly introduce a triple-level variable to capture the shared background among PQA. Then, two separate variables conditioned on the triple-level variable are used to represent the latent space for question and answer, and the question variable is also dependent on the answer one. During training, the latent variables are constrained to reconstruct both the original question and answer according to the hierarchical structure we define, making sure the triple-wise relationship flows through the latent variables without any loss. For the question generation process, we sample the triple-level and answer variable given a post, then obtain the question variable conditioned on them, and finally generate a question based on the post, triple-level and question variables. Experimental results on a large-scale CQG dataset show that GTM can generate more fluent, coherent, and intriguing questions for open-domain conversations.

The main contribution is threefold: 
\begin{itemize}
    \item To generate coherent and informative questions in the CQG task, we propose a generative triple-wise model that models the semantic relationship of a triple in three levels: PQA, PQ, and QA.
    
    \begin{figure}[!htb]
        \begin{center}
        \includegraphics[width=0.95\linewidth]{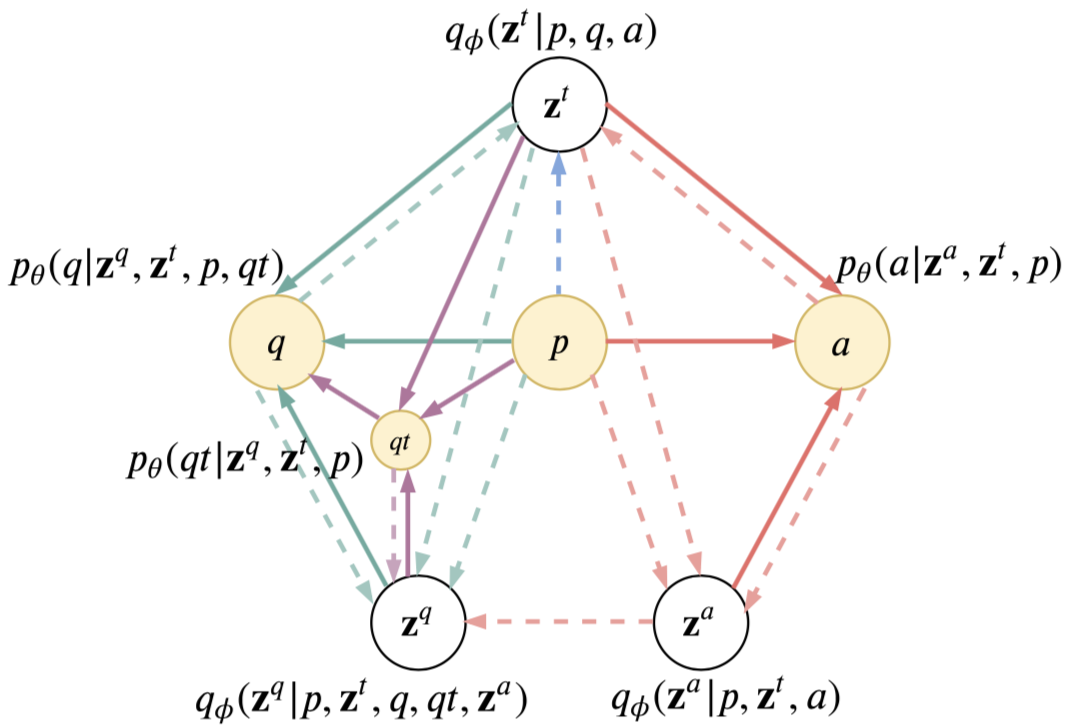}
        \end{center}
        \caption{The graphical representation of GTM for training process. ${\bf z}^{t}$ is used to capture the shared background among PQA, while ${\bf z}^q$ and ${\bf z}^a$ are used to model the diversity in PQ and QA pairs. Solid arrows illustrate the generation of ${\bm q}$, ${\bm a}$ (not used in inference), and $qt$, while dashed arrows are for posterior distributions of latent variables.}
        \label{fig:graphical}
    \end{figure}
    
    \item Our variational hierarchical structure can not only utilize the ``future'' information (answer), but also capture one-to-many mappings in PQ and QA, which matches the open-domain scenario well.
    \item Experimental results on a large-scale CQG corpus show that our method significantly outperforms the state-of-the-art baselines in both automatic and human evaluations.
\end{itemize}

\begin{figure*}[!htb]
\begin{center}
   \includegraphics[width=0.95\linewidth]{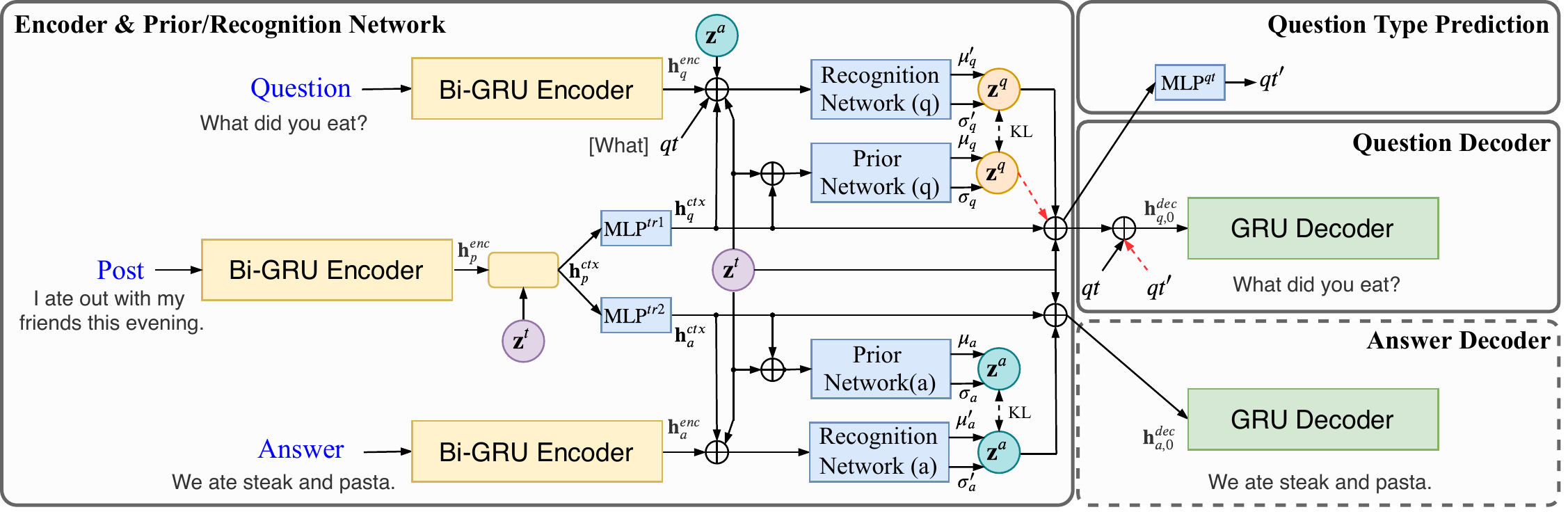}
\end{center}
   \caption{The architecture of GTM. $\oplus$ denotes the concatenation operation. In training process, latent variables obtained from recognition networks and the real question type $qt$ are used for decoding. Red dashed arrows refer to inference process, in which we get latent variables from prior networks, and the predicted question type $qt'$ is fed into the question decoder. The answer decoder is only utilized during training to assist the triple-wise modeling.}
\label{fig:overview}
\end{figure*}

\section{Proposed Model}
\label{sec:method}
Given a post as the input, the goal of CQG is to generate the corresponding question. Following the work of \citet{zhao2017learning} and \citet{wang2019answer}, we leverage the question type $qt$ to control the generated question, and take advantage of the answer information ${\bm a}$ to improve coherence. In training set, each conversation is represented as $\{{\bm p}, {\bm q}, qt, {\bm a}\}$, consisting of post ${\bm p} = {\{p_i\}_{i=1}^{|{\bm p}|}}$, question ${\bm q} = {\{q_i\}_{i=1}^{|{\bm q}|}}$ with its question type $qt$, and answer ${\bm a} = {\{a_i\}_{i=1 }^{|{\bm a}|}}$.

\subsection{Overview}
\label{sec:overview}
The graphical model of GTM for training process is shown in Figure \ref{fig:graphical}. $\theta$, $\varphi$, and $\phi$ are used to denote parameters of generation, prior, and recognition network, respectively. We integrate answer generation to assist question generation with hierarchical latent variables. Firstly, a triple-level variable ${\bf z}^t$ is imported to capture the shared background and is inferred from PQA utterances. Then answer latent variable ${\bf z}^a$ and question latent variable ${\bf z}^q$ are sampled from Gaussian distributions conditioned on both post and ${\bf z}^t$. To ensure that the question is controlled by answer, ${\bf z}^q$ is also dependent on ${\bf z}^a$.

\subsection{Input Representation}
We use a bidirectional GRU \cite{cholearning} as encoder to capture the semantic representation of each utterance. Take post ${\bm p}$ as an example. Each word in ${\bm p}$ is firstly encoded into its embedding vector. The GRU then computes forward hidden states $\{{\overrightarrow{h}_i}\}_{i=1}^{|{\bm p}|}$ and backward hidden states $\{{\overleftarrow{h}_i}\}_{i=1}^{|{\bm p}|}$:
\begin{align*}
&\overrightarrow{h}_i = \overrightarrow{\text{GRU}}(e_{p_i}, \overrightarrow{h}_{i-1}), \\
&\overleftarrow{h}_i =  \overleftarrow{\text{GRU}}(e_{p_i}, \overleftarrow{h}_{i+1}),
\end{align*}
where $e_{p_i}$ is employed to represent the embedding vector of word $p_i$. We finally get the post representation by concatenating the last hidden states of two directions ${\bf h}^{enc}_p = [{\overrightarrow{h}_{|{\bm p}|}};{\overleftarrow{h}_1}]$. Similarly, we can obtain representations of question ${\bm q}$ and answer ${\bm a}$, denoted as ${\bf h}^{enc}_q$ and ${\bf h}^{enc}_a$, respectively. 

The question type $qt$ is represented by a real-valued, low dimensional vector ${\bf v}_{qt}$ which is updated during training and is regarded as a linguistic feature that benefits the training of latent variables \cite{zhao2017learning}. We use the actual question type $qt$ during training to provide the information of interrogative words that is the most important feature to distinguish question types.

\subsection{Triple-level Latent Variable}
To capture the shared background of entire triple, we introduce a triple-level latent variable ${\bf z}^t$ that is inferred from PQA  utterances and is in turn responsible for generating the whole triple. Inspired by \citet{park2018hierarchical}, we use a standard Gaussian distribution as the prior distribution of ${\bf z}^t$:
\begin{equation*}
    p_\varphi({\bf z}^t) = \mathcal{N}({\bf z|0,I}), 
\end{equation*}
where ${\bf I}$ represents the identity matrix.

For the inference of ${\bf z}^t$ in training set, we consider three utterance representations ${\bf h}^{enc}_p$, ${\bf h}^{enc}_q$ and ${\bf h}^{enc}_a$ as a sequence, and use a bidirectional GRU to take individual representation as the input of each time step. The triple representation ${\bf h}^t$ is obtained by concatenating the last hidden states of both directions. Then, ${\bf z}^t$ is sampled from:
\begin{align*}
    & q_\phi({\bf z}^t|{\bm p}, {\bm q}, {\bm a}) = \mathcal{N}({\bf z}|\bm{\mu}^t, \bm{\sigma}^t{\bf I}), \\
    & \bm{\mu}^t = \text{MLP}^t_\phi({\bf h}^t), \\
    & \bm{\sigma}^t = \text{softplus}(\text{MLP}^t_\phi({\bf h}^t)),
\end{align*}
where MLP($\cdot$) is a feed-forward network, and softplus function is a smooth approximation to ReLU and can be used to ensure positiveness \cite{park2018hierarchical,serban2017hierarchical}.

\subsection{One-to-many Mappings}
After obtaining ${\bf z}^t$, we use a GRU $f$ to get a vector ${\bf h}^{ctx}_p$ for connecting ${\bm p}$ and ${\bm q}$/${\bm a}$. ${\bf h}^{ctx}_p$ is then transformed to ${\bf h}^{ctx}_q$ and ${\bf h}^{ctx}_a$ that are used in prior and recognition networks for ${\bf z}^q$ and ${\bf z}^a$:
\begin{align*}
    {\bf h}^{ctx}_p &= f({\bf z}^{t}, {\bf h}^{enc}_p), \\
    {\bf h}^{ctx}_q &= \text{MLP}_\theta^{tr1}({\bf h}^{ctx}_p), \\
    {\bf h}^{ctx}_a &= \text{MLP}_\theta^{tr2}({\bf h}^{ctx}_p).
\end{align*}

To model one-to-many mappings in PQ and QA pairs under the control of ${\bf z}^t$, we design two utterance-level variables, ${\bf z}^q$ and ${\bf z}^a$, to represent latent spaces of question and answer. We define the prior and posterior distributions of ${\bf z}^a$ as follows:
\begin{align*}
    &p_\varphi({\bf z}^a|{\bm p}, {\bf z}^t) = \mathcal{N}({\bf z}|\bm{\mu}_a, \bm{\sigma}_a{\bf I}), \\
    &q_\phi({\bf z}^a|{\bm p}, {\bf z}^t, {\bm a}) = \mathcal{N}({\bf z}|\bm{\mu}^{'}_a, \bm{\sigma}^{'}_a{\bf I}),
\end{align*}
where $\bm{\mu}_{a}$, $\bm{\sigma}_{a}$, $\bm{\mu}^{'}_{a}$, and $\bm{\sigma}^{'}_{a}$, the parameters of two Gaussian distributions, are calculated as:
\begin{align*}
    \bm{\mu}_{a} &= \text{MLP}^a_\varphi([{\bf h}^{ctx}_a; {\bf z}^t]), \\
    \bm{\sigma}_{a} &= \text{softplus}(\text{MLP}^a_\varphi([{\bf h}^{ctx}_a; {\bf z}^t])), \\
    \bm{\mu}^{'}_{a} &= \text{MLP}^a_\phi([{\bf h}^{ctx}_a; {\bf z}^t; {\bf h}^{enc}_a]), \\
    \bm{\sigma}^{'}_{a} &= \text{softplus}(\text{MLP}^a_\phi([{\bf h}^{ctx}_a; {\bf z}^t; {\bf h}^{enc}_a])).
\end{align*}

To make sure the content of question is also decided by answer and improve their relatedness, we import ${\bf z}^a$ into ${\bf z}^q$ space. The prior and posterior distributions of ${\bf z}^q$ are computed as follows:
\begin{align*}
    &p_\varphi({\bf z}^q|{\bm p}, {\bf z}^t, {\bf z}^a) = \mathcal{N}({\bf z}|\bm{\mu}_q, \bm{\sigma}_q{\bf I}), \\
    &q_\phi({\bf z}^q|{\bm p}, {\bf z}^t, {\bm q}, qt, {\bf z}^a) = \mathcal{N}({\bf z}|\bm{\mu}^{'}_q, \bm{\sigma}^{'}_q{\bf I}),
\end{align*}
where $\bm{\mu}_{q}$, $\bm{\sigma}_{q}$, $\bm{\mu}^{'}_{q}$, and $\bm{\sigma}^{'}_{q}$ are calculated as:
\begin{align*}
    \bm{\mu}_{q} &= \text{MLP}^q_\varphi([{\bf h}^{ctx}_q; {\bf z}^t; {\bf z}^a]), \\
    \bm{\sigma}_{q} &= \text{softplus}(\text{MLP}^q_\varphi([{\bf h}^{ctx}_q; {\bf z}^t; {\bf z}^a])), \\
    \bm{\mu}^{'}_{q} &= \text{MLP}^q_\phi([{\bf h}^{ctx}_q; {\bf z}^t; {\bf h}^{enc}_q; {\bf v}_{qt}; {\bf z}^a]), \\
    \bm{\sigma}^{'}_{q} &= \text{softplus}(\text{MLP}^q_\phi([{\bf h}^{ctx}_q; {\bf z}^t; {\bf h}^{enc}_q; {\bf v}_{qt}, {\bf z}^a])). 
\end{align*}

\subsection{Question Generation Network}
Following the work of \citet{zhao2017learning} and \citet{wang2019answer}, a question type prediction network $\text{MLP}^{qt}$ is introduced to approximate $p_\theta(qt|{\bf z}^q, {\bf z}^t, {\bm p})$ in training process and produces question type $qt'$ during inference.

As shown in Figure \ref{fig:overview},  there are two decoders in our model, one is for answer generation that is an auxiliary task and only exists in the training process, and the other is for desired question generation. The question decoder employs a variant of GRU that takes the concatenation result of ${\bf z}^q$, ${\bf z}^t$, ${\bf h}^{ctx}_q$, and $qt$ as initial state ${\bf s}_0$, i.e., ${\bf s}_0 = [{\bf z}^q; {\bf z}^t, {\bf h}^{ctx}_q, qt]$. For each time step $j$, it calculates the context vector ${\bf c}_j$ following \citet{bahdanau2015neural}, and computes
the probability distribution $p_\theta({\bm q}|{\bf z}^q, {\bf z}^t, {\bm p}, qt)$ over all words in the vocabulary: 
\begin{align*}
    &{\bf s}_j = \text{GRU}({\bf e}_{j-1}, {\bf s}_{j-1}, {\bf c}_j) \\
    &{\bf \Tilde{s}}_j = \text{MLP}([{\bf e}_{j-1};{\bf c}_j;{\bf s}_j]), \\
    & p_\theta({\bm q}_j|{\bm q}_{<j}, {\bf z}^q, {\bf z}^t, {\bm p}, qt)= \text{softmax}({\bf W_o}{\bf \Tilde{s}}_j),
\end{align*}
where ${\bf e}_{j-1}$ represents the embedding vector of the $(j-1)$-th question word. Similarly, the answer decoder receives the concatenation result of ${\bf z}^a$, ${\bf z}^t$, and ${\bf h}^{ctx}_a$ as initial state to approximate the probability $p_\theta({\bm a}|{\bf z}^a, {\bf z}^t, {\bm p})$.

\subsection{Training and Inference}
Importantly, our model GTM is trained to maximize the log-likelihood of the joint probability $p({\bm p},{\bm q}, {\bm a}, qt)$:
\begin{equation*}
    \mathrm{log} p({\bm p},{\bm q}, {\bm a}, qt) = \mathrm{log}\int_{{\bf z}^t}p({\bm p},{\bm q}, {\bm a}, qt, {\bf z}^t).
\end{equation*}

However, the optimization function is not directly tractable. Inspired by \citet{serban2017hierarchical} and \citet{park2018hierarchical}, we convert it to the following objective that is based on the evidence lower bound and needs to be maximized in training process:
\begin{equation*}
\begin{aligned}
    &\mathcal{L}_{\mathrm{GTM}} = \\ 
    & -KL(q_\phi({\bf z}^t|{\bm p}, {\bm q}, {\bm a})||p_\varphi({\bf z}^t)) \\
    & -KL(q_\phi({\bf z}^a|{\bm p}, {\bf z}^t, {\bm a})||p_\varphi({\bf z}^a|{\bm p}, {\bf z}^t)) \\
    & -KL(q_\phi({\bf z}^q|{\bm p}, {\bf z}^t, {\bm q}, qt, {\bf z}^a)||p_\varphi({\bf z}^q|{\bm p}, {\bf z}^t, {\bf z}^a)) \\
    & +\mathbb{E}_{{\bf z}^a, {\bf z}^t \sim q_\phi}[\log p_\theta({\bm a}|{\bf z}^a, {\bf z}^t, {\bm p})] \\
    & +\mathbb{E}_{{\bf z}^q, {\bf z}^t \sim q_\phi}[\log p_\theta({\bm q}|{\bf z}^q, {\bf z}^t, {\bm p}, qt)] \\
    & +\mathbb{E}_{{\bf z}^q, {\bf z}^t \sim q_\phi}[\log p_\theta(qt|{\bf z}^q, {\bf z}^t, {\bm p})].
\label{eq:obj}
\end{aligned}
\end{equation*}

The objective consists of two parts: the variational lower bound (the first five lines) and question type prediction accuracy (the last line). Meanwhile, the variational lower bound includes the reconstruction terms and KL divergence terms based on three hierarchical latent variables. The gradients to the prior and recognition networks can be estimated using the reparameterization trick \cite{kingma2013auto}.

During inference, latent variables obtained via prior networks and predicted question type $qt'$ are fed to the question decoder, which corresponds to red dashed arrows in Figure \ref{fig:overview}. The inference process is as follows: 

(1) Sample triple-level LV: ${\bf z}^t \sim q_\phi({\bf z}^t|{\bm p})$\footnote{Inspired by \citet{park2018hierarchical}, using ${\bf z}^t$ inferred from post with the posterior distribution is better than sampling it from the prior one, i.e., a standard Gaussian distribution.}. 

(2) Sample answer LV: ${\bf z}^a \sim p_\varphi({\bf z}^a|{\bm p}, {\bf z}^t)$. 

(3) Sample question LV: ${\bf z}^q \sim p_\varphi({\bf z}^q|{\bm p}, {\bf z}^t, {\bf z}^a)$.

(4) Predict question type: $qt \sim p_\theta(qt|{\bf z}^q, {\bf z}^t, {\bm p})$. 

(5) Generate question: ${\bm q} \sim p_\theta({\bf z}^q, {\bf z}^t, {\bm p}, qt)$. 

\section{Experiments}
\label{sec:experiments}
In this section, we conduct experiments to evaluate our proposed method. We first introduce some empirical settings, including dataset, hyper-parameters, baselines, and evaluation measures. Then we illustrate our results under both automatic and human evaluations. Finally, we give out some cases generated by different models and do further analyses over our method.

\subsection{Dataset}
We apply our model on a large-scale CQG corpus\footnote{\url{https://drive.google.com/drive/folder/1wNG30YPHiMc_ZNyE3BH5wa1uVtR8l1pG}} extracted from Reddit\footnote{\url{http://www.reddit.com}} by \citet{wang2019answer}. There are over 1.2 million PQA triples which have been divided into training/validation/test set with the number of 1,164,345/30,000/30,000. The dataset has been tokenized into words using the NLTK tokenizer \cite{bird2009natural}. The average number of words in post/question/answer is 18.84/19.03/19.30, respectively. Following \citet{fan2018question} and \citet{wang2019answer}, we categorize questions in training and validation set into 9 types based on interrogative words, i.e., ``what'', ``when'', ``where'', ``who'', ``why'', ``how'', ``can (could)'', ``do (did, does)'', ``is (am, are, was, were)'' 

\subsection{Hyper-parameter Settings}
We keep the top 40,000 frequent words as the vocabulary and the sentence padding length is set to 30. The dimension of GRU layer, word embedding and latent variables is 300, 300, and 100. The prior networks and MLPs have one hidden layer with size 300 and tanh non-linearity, while the number of hidden layers in recognition networks for both triple-level and utterance-level variables is 2. We apply dropout ratio of 0.2 during training. The mini-batch size is 64. For optimization, we use Adam \cite{kingma2014adam} with a learning rate of 1e-4. In order to alleviate degeneration problem of variational framework \cite{park2018hierarchical}, we apply KL annealing, word drop \cite{bowman2015generating} and bag-of-word (BOW) loss \cite{zhao2017learning}\footnote{The total BOW loss is calculated as the sum of all BOW losses between each latent variable and ${\bm q}$/${\bm a}$. Please refer to \citet{park2018hierarchical} for more details.}. The KL multiplier $\lambda$ gradually increases from 0 to 1, and the word drop probability is 0.25. We use Pytorch to implement our model, and the model is trained on Titan Xp GPUs.

\begin{table*}[!htb]
    \small
    \centering
    \begin{tabular}{l|ccc|cc|cc|cc}
    \toprule
        \multirow{2}{*}{Model} & \multicolumn{3}{c|}{Embedding Metrics} & \multicolumn{2}{c|}{Diversity} & \multicolumn{2}{c|}{BLEU Scores} & \multicolumn{2}{c}{RUBER Scores} \\ \cline{2-10}
        ~ & Average & Extrema & Greedy & Dist-1 & Dist-2 & BLEU-1 & BLEU-2 & RubG & RubA \\ \hline
        S2S-Attn & 0.634 & 0.322 & 0.413 & 0.0132 & 0.0830 & 0.0936 & 0.0298 & 0.584 & 0.622 \\ 
        CVAE & 0.646 & 0.337 & 0.421 & 0.0160 & 0.1599 & 0.1422 & 0.0306 & 0.649 & 0.687 \\
        kgCVAE & 0.647 & 0.332 & 0.425 & 0.0153 & 0.1587 & 0.1491 & 0.0310 & 0.650 & 0.682 \\ 
        STD & 0.637 & 0.326 & 0.418 & 0.0144 & 0.1325 & 0.1327 & 0.0302 & 0.633 & 0.663\\
        HTD & 0.648 & 0.330 & 0.423 & 0.0154 & 0.1582 & 0.1475 & 0.0314 & 0.653 & 0.689\\ 
        RL-CVAE & 0.662 & 0.343 & 0.437 & 0.0161 & 0.1785 & 0.1503 & 0.0320 & 0.660 & 0.701 \\ \hline
        GTM-${\bf z}^t$ & 0.672 & 0.351 & 0.448 & 0.0165 & 0.1872 & 0.1521 & {\bf 0.0332} & 0.661 & 0.710 \\ 
        GTM-a & 0.653 & 0.338 & 0.428 & 0.0158 & 0.1679 & 0.1482 & 0.0317 & 0.657 & 0.692 \\
        GTM-${\bf z}^q/{\bf z}^a$ & 0.687 & 0.360 & 0.449 & 0.0170 & 0.1934 & 0.1528 & 0.0329 & 0.669 & 0.713 \\
        GTM & {\bf 0.697} & {\bf 0.365} & {\bf 0.454} & {\bf 0.0176} & {\bf 0.2028} & {\bf 0.1537} & 0.0331 & {\bf 0.671} & {\bf 0.720}\\
    \bottomrule
    \end{tabular}
    \caption{Automatic evaluation results for different models based on four types of metrics.}
    \label{tab:autoresults}
\end{table*}

\subsection{Baselines}
We compare our methods with four groups of representative models: (1) \textbf{S2S-Attn}: A simple Seq2Seq model with attention mechanism \cite{shang2015neural}. (2) \textbf{CVAE\&kgCVAE}: The CVAE model integrates an extra BOW loss to generate diverse questions. The kgCVAE is a knowledge-guided CVAE that utilizes some linguistic cues (question types in our experiments) to learn meaningful latent variables \cite{zhao2017learning}. (3) \textbf{STD\&HTD}: The STD uses soft typed decoder that estimates a type distribution over word types, and the HTD uses hard typed decoder that specifies the type of each word explicitly with Gumbel-softmax \cite{wang2018learning}. (4) \textbf{RL-CVAE}: A reinforcement learning method that regards the coherence score (computed by a one-to-one matching network) of a pair of generated question and answer as the reward function \cite{wang2019answer}. RL-CVAE is the first work to utilize the future information, i.e., answer, and is also the state-of-the-art model for CQG\footnote{For those methods with open-source codes, we run the original codes; otherwise, we re-implement them based on the corresponding paper.}. 

Additionally, we also conduct ablation study to better analyze our method as follows: (5) \textbf{GTM-${\bf z}^t$}: GTM without the triple-level latent variable, which means ${\bf z}^t$ is not included in the prior and posterior distributions of both ${\bf z}^p$ and ${\bf z}^a$. (6) \textbf{GTM-a}: the variant of GTM that does not take answer into account. That is, answer decoder and ${\bf z}^a$ are removed from the loss function and the prior and posterior distributions of ${\bf z}^q$. Besides, ${\bf z}^t$ here does not capture the semantics from answer. (7) \textbf{GTM-${\bf z}^q/{\bf z}^a$}: GTM variant in which distributions of ${\bf z}^q$ are not conditioned on ${\bf z}^a$, i.e., the fact that the content of question is also controlled by answer is not modelled explicitly by latent variables.

In our model, we use an MLP to predict question types during inference, which is different from the conditional training (CT) methods \cite{li2016persona,zhou2018emotional,shen2020cdl} that provide the controllable feature, i.e., question types, in advance for inference. Therefore, we do not consider CT-based models as comparable ones.

\subsection{Evaluation Measures}
To better evaluate our results, we use both quantitative metrics and human judgements in our experiments.

\subsubsection*{Automatic Metrics}
\label{sec:autometrics}
For automatic evaluation, we mainly choose four kinds of metrics:
(1) \textbf{BLEU Scores:} BLEU \cite{papineni2002bleu} calculates the n-gram overlap score of generated questions against ground-truth questions. We use BLEU-1 and BLEU-2 here and normalize them to 0 to 1 scale. (2) \textbf{Embedding Metrics:} Average, Greedy and Extrema metrics are embedding-based and measure the semantic similarity between the words in generated questions and ground-truth questions \cite{serban2017hierarchical, liu2016not}. We use word2vec embeddings trained on the Google News Corpus\footnote{\url{https://code.google.com/archive/p/word2vec/}} in this part. Please refer to \citet{serban2017hierarchical} for more details. (3) \textbf{Dist-1\& Dist-2:} Following the work of \citet{li2016diversity}, we apply {\it Distinct} to report the degree of diversity. {\it Dist-1/2} is defined as the ratio of unique uni/bi-grams over all uni/bi-grams in generated questions. (4) \textbf{RUBER Scores:} Referenced metric and Unreferenced metric Blended Evaluation Routine \cite{tao2018ruber} has shown a high correlation with human annotation in open-domain conversation evaluation. There are two versions, one is RubG based on geometric averaging and the other is RubA based on arithmetic averaging.

Embedding metrics and BLEU scores are used to measure the similarity between generated and ground-truth questions. RubG/A reflects the semantic coherence of PQ pairs \cite{wang2019answer}, while Dist-1/2 evaluates the diversity of questions.

\subsubsection*{Human Evaluation Settings}
Inspired by \citet{wang2019answer}, \citet{shen2019modeling}, and \citet{wang2018learning}, we use following three criteria for human evaluation: (1) {\bf Fluency} measures whether the generated question is reasonable in logic and grammatically correct. (2) {\bf Coherence} denotes whether the generated question is semantically consistent with the given post. Incoherent questions include dull cases. (3) {\bf Willingness} measures whether a user is willing to answer the question. This criterion is to justify how likely the generated questions can elicit further interactions.

We randomly sample 500 examples from test set, and generate questions using models mentioned above. Then, we send each post and corresponding 10 generated responses to three human annotators without order, and require them to evaluate whether each question satisfies criteria defined above. All annotators are postgraduate students and not involved in other parts of our experiments.

\subsection{Experimental Results}
Now we demonstrate our experimental results on both automatic evaluation and human evaluation.

\subsubsection*{Automatic Evaluation Results}
Now we demonstrate our experimental results on both automatic evaluation and human evaluation.
The automatic results are shown in Table \ref{tab:autoresults}. The top part is the results of all baseline models, and we can see that GTM outperforms other methods on all metrics (significance tests \cite{koehn2004statistical}, $p$-value $<$ 0.05), which indicates that our proposed model can improve the overall quality of generated questions. Specifically, Dist-2 and RubA have been improved by 2.43\% and 1.90\%, respectively, compared to the state-of-the-art RL-CVAE model. First, higher embedding metrics and BLEU scores show that questions generated by our model are similar to ground truths in both topics and contents. Second, taking answer into account and using it to decide the expression of question can improve the consistency of PQ pairs evaluated by RUBER scores. Third, higher distinct values illustrate that one-to-many mappings in PQ and QA pairs make the generated responses more diverse. 

The bottom part of Table \ref{tab:autoresults} shows the results of our ablation study, which demonstrates that taking advantage of answer information, modeling the shared background in entire triple, and considering one-to-many mappings in both PQ and QA pairs can help enhance the performance of our hierarchical variational model in terms of relevance, coherence and diversity.

\subsubsection*{Human Evaluation Results}
As shown in Table \ref{tab:humanresults}, GTM can alleviate the problem of generating dull and deviated questions compared with other models (significance tests \cite{koehn2004statistical}, $p$-value $<$ 0.05). Both our proposed model and the state-of-the-art model RL-CVAE utilize the answer information and the results of them could prove that answers assist the question generation process. Besides, GTM can produce more relevant and intriguing questions, which indicates the effectiveness of modeling the shared background and one-to-many mappings in CQG task. The inter-annotator agreement is calculated with the Fleiss' kappa \cite{fleiss1973equivalence}. Fleiss’ kappa for Fluency, Coherence and Willingness is 0.493, 0.446 and 0.512, respectively, indicating ``Moderate Agreement'' for all three criteria.

\subsection{Question-Answer Coherence Evaluation}

Automatic metrics in Section ``Automatic Metrics'' are designed to compare generated questions with ground-truth ones (RUBER also takes the post information into consideration), but ignore answers in the evaluation process. To measure the semantic coherence between generated questions and answers, we apply two methods \cite{wang2019answer}: (1) Cosine Similarity: We use the pre-trained Infersent model\footnote{The Infersent model is trained to predict the meaning of sentences based on natural language inference, and the cosine similarity computed with it is more consistent with human's judgements, which performs better than the pre-trained Transformer/BERT model in our experiments.} \cite{conneau2017supervised} to obtain sentence embeddings and calculate cosine similarity between the embeddings of generated responses and answers. (2) Matching Score: We use the GRU-MatchPyramid \cite{wang2019answer} model that adds\begin{table}[htb]
    \small
    \centering
    \begin{tabular}{l|ccc}
    \toprule
        Model & Fluency & Coherence & Willingness \\ \hline
        S2S-Attn & 0.482 & 0.216 & 0.186 \\
        CVAE & 0.462 & 0.484 & 0.428 \\
        kgCVAE & 0.474 & 0.536 & 0.476 \\
        STD & 0.488 & 0.356 & 0.286 \\ 
        HTD & 0.526 & 0.504 & 0.414 \\ 
        RL-CVAE & 0.534 & 0.578 & 0.508 \\ \hline
        GTM-${\bf z}^t$ & 0.538 & 0.580 & 0.516 \\ 
        GTM-a & 0.532 & 0.570 & 0.512 \\
        GTM-${\bf z}^q/{\bf z}^a$ & 0.542 & 0.586 & 0.520 \\
        GTM & {\bf 0.548} & {\bf 0.608} & {\bf 0.526} \\
    \bottomrule
    \end{tabular}
    \caption{Results for human evaluation.}
    \label{tab:humanresults}
\end{table}\begin{table}[htb]
    \small
    \centering
    \begin{tabular}{l|cc}
    \toprule
        Model & Cosine Similarity & Matching Score \\ \hline
        S2S-Attn & 0.498 & 5.306 \\
        CVAE & 0.564 & 8.047 \\
        kgCVAE & 0.578 & 8.054 \\
        STD & 0.542 & 6.879 \\ 
        HTD & 0.583 & 8.059 \\ 
        RL-CVAE & 0.607 & 8.423 \\ \hline
        GTM-${\bf z}^t$ & 0.613 & 8.427 \\
        GTM-a & 0.605 & 8.424 \\
        GTM-${\bf z}^q/{\bf z}^a$ & 0.618 & 8.472 \\
        GTM & {\bf 0.629} & {\bf 8.517} \\
    \bottomrule
    \end{tabular}
    \caption{Evaluation results for QA coherence.}
    \label{tab:q-aresults}
\end{table} the MatchPyramid network \cite{pang2016text} on top of a bidirectional GRU to calculate the semantic coherence. As shown in Table \ref{tab:q-aresults}, questions generated by GTM are more coherent to answers. Attributing to the design of triple-level latent variable that captures the shared background, one-to-many mappings in PQ and QA pairs, and relationship modeling for ${\bf z}^q$ and ${\bf z}^a$, GTM can improve the relevance in QA pairs.

\begin{table*}[!htb]
    \small
    \centering
    \begin{tabular}{p{3.5cm}|p{7cm}|p{3.5cm}}
    \toprule
        \multicolumn{1}{m{3.5cm}<{\centering}|}{\bf Post} &  \multicolumn{1}{m{7cm}<{\centering}|}{\bf Question} & \multicolumn{1}{m{3.5cm}<{\centering}}{\bf Answer} \\ \hline
        & S2S-Attn: what does that mean? & \\ 
        & CVAE: what's the best way to do that? & because i like the aspect of \\
        and it's a donation to food & kgCVAE: is there anything else to buy? & song criticising may \\
        banks, even if you dislike & STD: is it a good donation? & reaching to the charts as \\
        the song. & HTD: where are those food banks? & well as the fact it goes to \\
        & RL-CVAE: why do we need to do the stupid thing? & charity. \\
        & {\bf GTM: why not just donate money to food banks?} \\ \hline
        & S2S-Attn: what colors do you have? & \\
        & CVAE: are you colorblind? & i have some coming, \\
        it's a very blue pen. i'm & kgCVAE: it has nothing to complain. where's the pen? & hopefully it will be here \\
        more into brighter colours, & STD: what color are you using? & early next week. right now \\
        but this is still pretty nice. & HTD: what's the colour scheme? & it's got green ink and it's \\
        & RL-CVAE: what kind of ink should i buy? & really messing me up. \\
        & {\bf GTM: i'll take it though. do you also sell the ink?} &  \\
    \bottomrule
    \hline
    \end{tabular}
    \caption{Two cases comparison among GTM and other baselines.}
    \label{tab:case_study}
\end{table*}

\subsection{Case Study}
In Table \ref{tab:case_study}, we list the generated results of two posts from the test set to compare the performance of different models. 

In the first case, both the post and answer mention two topics, ``donation'' and ``song'', so the question is better to consider their relations. Besides, the answer here begins with ``because'', then ``why'' and ``what (reason)'' questions are reasonable. For the second case, the post only talks about ``pen'', while the answer refers to ``ink'', which means there is a topic transition the question needs to cover. The second case shows the effectiveness of an answer that not only decides the expression of question but also improves the entire coherence of a tripe. Questions generated by GTM are more relevant to both posts and answers, and could attract people\begin{figure}[!htb]
\begin{center}
   \includegraphics[width=1.0\linewidth]{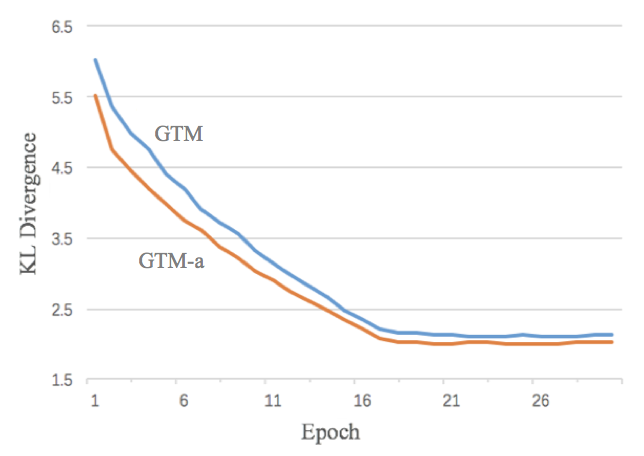}
\end{center}
   \caption{Total KL divergence (per word) of all latent variables in GTM and GTM-a model (first 30 epochs of validation set).}
\label{fig:kl}
\end{figure} to give an answer to them. However, other baselines may generate dull or deviated responses, even the RL-CVAE model that considers the answer information would only contain the topic words in answers (e.g., the question in case two), but fail to ensure the PQA coherence.

\subsection{Further Analysis of GTM}
Variational models suffer from the notorious degeneration problem, where the decoders ignore latent variables and reduce to vanilla Seq2Seq models \cite{zhao2017learning,park2018hierarchical,wang2019answer}. Generally, KL divergence measures the amount of information encoded in a latent variable. In the extreme case where the KL divergence of latent variable ${\bf z}$ equals to zero, the model completely ignores ${\bf z}$, i.e., it degenerates. Figure \ref{fig:kl} shows that the total KL divergence of GTM model maintains around 2 after 18 epochs indicating that the degeneration problem does not exist in our model and latent variables can play their corresponding roles.

\section{Related Work}
\label{sec:relatedwork}
The researches on open-domain dialogue systems have developed rapidly \cite{majumder2020,zhan-etal-2021-augmenting,shen2021learning}, and our work mainly touches two fields: open-domain conversational question generation (CQG), and context modeling in dialogue systems. We introduce these two fields as follows and point out the main differences between our method and previous ones.

\subsection{CQG}
\label{rel:cqg}
Traditional question generation (TQG) has been widely studied and can be seen in reading comprehension \cite{zhou2019question,kim2019improving}, sentence transformation \cite{vanderwende2008importance}, question answering \cite{li2019improving,nema2019let}, visual question generation \cite{fan2018question} and task-oriented dialogues \cite{li2016learning}. In such tasks, finding information via a generated question is the major goal and the answer is usually part of the input. Different from TQG, CQG aims to enhance the interactiveness and persistence of conversations \cite{wang2018learning}. Meanwhile, the answer is the ``future'' information which means it is unavailable in the inference process. \citet{wang2018learning} first studied on CQG, and they used soft and hard typed decoders to capture the distribution of different word types in a question. \citet{hu2018aspect} added a target aspect in the input and proposed an extended Seq2Seq model to generate aspect-specific questions. \citet{wang2019answer} devised two methods based on either reinforcement learning or generative adversarial network (GAN) to further enhance semantic coherence between posts and questions under the guidance of answers.
 
\subsection{Context Modeling in Dialogue Systems}
\label{rel:cont}
Existing methods mainly focus on the historical context in multi-turn conversations, and hierarchical models occupy a vital position in this field. \citet{serban2016building} proposed the hierarchical recurrent encoder-decoder (HRED) model with a context RNN to integrate historical information from utterance RNNs. To capture utterance-level variations, \citet{serban2017hierarchical} raised a new model Variational HRED (VHRED) that augments HRED with CVAEs. After that, VHCR \cite{park2018hierarchical} added a conversation-level latent variable on top of the VHRED, while CSRR \cite{shen2019modeling} used three-hierarchy latent variables to model the complex dependency among utterances. In order to detect relative utterances in context, \citet{tian2017make} and \citet{zhang2018context} applied cosine similarity and attention mechanism, respectively. HRAN \cite{xing2018hierarchical} combined the attention results on both word-level and utterance-level. Besides, the future information has also been considered for context modeling. \citet{shen2018nexus} separated the context into history and future parts, and assumed that each of them conditioned on a latent variable is under a Gaussian distribution. \citet{feng2020posterior} used future utterances in the discriminator of a GAN, which is similar to \citet{wang2019answer}.

The differences between our method and aforementioned ones in Section \ref{rel:cqg} and \ref{rel:cont} are: (1) Rather than dividing PQA triples into two parts, i.e., PQ (history and current utterances) and QA (current and future utterances) pairs, we model the entire coherence by utilizing a latent variable to capture the share background in a triple. (2) Instead of regarding the relationship between question and answer as a text matching task that lacks the consideration of diversity, we incorporate utterance-level latent variables to help model one-to-many mappings in both PQ and QA pairs.

\section{Conclusion}
\label{sec:conclusion}
We propose a generative triple-wise model for generating appropriate questions in open-domain conversations, named GTM. GTM models the entire background in a triple and one-to-many mappings in PQ and QA pairs simultaneously with latent variables in three hierarchies. It is trained in a one-stage end-to-end framework without pre-training like the previous state-of-the-art model that also takes answer into consideration. Experimental results on a large-scale CQG dataset show that GTM can generate fluent, coherent, informative as well as intriguing questions.

\section*{Acknowledgements}
We would like to thank all the reviewers for their insightful and valuable comments and suggestions.

\bibliographystyle{acl_natbib}
\bibliography{anthology,acl2021}

\begin{thebibliography}{45}
\expandafter\ifx\csname natexlab\endcsname\relax\def\natexlab#1{#1}\fi

\bibitem[{Bahdanau et~al.(2015)Bahdanau, Cho, and Bengio}]{bahdanau2015neural}
Dzmitry Bahdanau, Kyunghyun Cho, and Yoshua Bengio. 2015.
\newblock Neural machine translation by jointly learning to align and
  translate.
\newblock In \emph{3rd International Conference on Learning Representations,
  ICLR 2015}.

\bibitem[{Bird et~al.(2009)Bird, Klein, and Loper}]{bird2009natural}
Steven Bird, Ewan Klein, and Edward Loper. 2009.
\newblock \emph{Natural language processing with Python: analyzing text with
  the natural language toolkit}.
\newblock " O'Reilly Media, Inc.".

\bibitem[{Bowman et~al.(2016)Bowman, Vilnis, Vinyals, Dai, Jozefowicz, and
  Bengio}]{bowman2015generating}
Samuel~R Bowman, Luke Vilnis, Oriol Vinyals, Andrew Dai, Rafal Jozefowicz, and
  Samy Bengio. 2016.
\newblock Generating sentences from a continuous space.
\newblock In \emph{Proceedings of the 20th SIGNLL Conference on Computational
  Natural Language Learning}, pages 10--21.

\bibitem[{Chen et~al.(2019)Chen, Peng, Wang, Xu, and Wu}]{chen2019generating}
Chaotao Chen, Jinhua Peng, Fan Wang, Jun Xu, and Hua Wu. 2019.
\newblock Generating multiple diverse responses with multi-mapping and
  posterior mapping selection.
\newblock In \emph{Proceedings of the 28th International Joint Conference on
  Artificial Intelligence}, pages 4918--4924. AAAI Press.

\bibitem[{Cho et~al.(2014)Cho, Gulcehre, Bahdanau, Schwenk, and
  Bengio}]{cholearning}
Kyunghyun Cho, Bart van Merri{\"e}nboer~Caglar Gulcehre, Dzmitry Bahdanau,
  Fethi Bougares~Holger Schwenk, and Yoshua Bengio. 2014.
\newblock Learning phrase representations using rnn encoder--decoder for
  statistical machine translation.
\newblock In \emph{Proceedings of the 2014 Conference on Empirical Methods in
  Natural Language Processing}, pages 1724--1734.

\bibitem[{Conneau et~al.(2017)Conneau, Kiela, Schwenk, Barrault, and
  Bordes}]{conneau2017supervised}
Alexis Conneau, Douwe Kiela, Holger Schwenk, Lo{\"\i}c Barrault, and Antoine
  Bordes. 2017.
\newblock \href {https://doi.org/10.18653/v1/D17-1070} {Supervised learning of
  universal sentence representations from natural language inference data}.
\newblock In \emph{Proceedings of the 2017 Conference on Empirical Methods in
  Natural Language Processing}, pages 670--680, Copenhagen, Denmark.
  Association for Computational Linguistics.

\bibitem[{Fan et~al.(2018)Fan, Wei, Li, Lan, and Huang}]{fan2018question}
Zhihao Fan, Zhongyu Wei, Piji Li, Yanyan Lan, and Xuanjing Huang. 2018.
\newblock A question type driven framework to diversify visual question
  generation.
\newblock In \emph{Proceedings of the 27th International Joint Conference on
  Artificial Intelligence}, pages 4048--4054. AAAI Press.

\bibitem[{Feng et~al.(2020)Feng, Chen, Li, and Yin}]{feng2020posterior}
Shaoxiong Feng, Hongshen Chen, Kan Li, and Dawei Yin. 2020.
\newblock Posterior-gan: Towards informative and coherent response generation
  with posterior generative adversarial network.
\newblock In \emph{Proceedings of the AAAI Conference on Artificial
  Intelligence}, volume~34, pages 7708--7715.

\bibitem[{Fleiss and Cohen(1973)}]{fleiss1973equivalence}
Joseph~L Fleiss and Jacob Cohen. 1973.
\newblock The equivalence of weighted kappa and the intraclass correlation
  coefficient as measures of reliability.
\newblock \emph{Educational and psychological measurement}, 33(3):613--619.

\bibitem[{Gao et~al.(2019)Gao, Lee, Zhang, Brockett, Galley, Gao, and
  Dolan}]{gao2019jointly}
Xiang Gao, Sungjin Lee, Yizhe Zhang, Chris Brockett, Michel Galley, Jianfeng
  Gao, and Bill Dolan. 2019.
\newblock \href {https://doi.org/10.18653/v1/N19-1125} {Jointly optimizing
  diversity and relevance in neural response generation}.
\newblock In \emph{Proceedings of the 2019 Conference of the North {A}merican
  Chapter of the Association for Computational Linguistics: Human Language
  Technologies, Volume 1 (Long and Short Papers)}, pages 1229--1238,
  Minneapolis, Minnesota. Association for Computational Linguistics.

\bibitem[{Hu et~al.(2018)Hu, Liu, Ma, Zhao, and Yan}]{hu2018aspect}
Wenpeng Hu, Bing Liu, Jinwen Ma, Dongyan Zhao, and Rui Yan. 2018.
\newblock Aspect-based question generation.

\bibitem[{Kim et~al.(2019)Kim, Lee, Shin, and Jung}]{kim2019improving}
Yanghoon Kim, Hwanhee Lee, Joongbo Shin, and Kyomin Jung. 2019.
\newblock Improving neural question generation using answer separation.
\newblock In \emph{Proceedings of the AAAI Conference on Artificial
  Intelligence}, volume~33, pages 6602--6609.

\bibitem[{Kingma and Ba(2015)}]{kingma2014adam}
Diederick~P Kingma and Jimmy Ba. 2015.
\newblock Adam: A method for stochastic optimization.
\newblock In \emph{the 3rd International Conference on Learning
  Representations}.

\bibitem[{Kingma and Welling(2014)}]{kingma2013auto}
Diederik~P Kingma and Max Welling. 2014.
\newblock Auto-encoding variational bayes.
\newblock In \emph{the 2nd International Conference on Learning
  Representations}.

\bibitem[{Koehn(2004)}]{koehn2004statistical}
Philipp Koehn. 2004.
\newblock \href {https://www.aclweb.org/anthology/W04-3250} {Statistical
  significance tests for machine translation evaluation}.
\newblock In \emph{Proceedings of the 2004 Conference on Empirical Methods in
  Natural Language Processing}, pages 388--395, Barcelona, Spain. Association
  for Computational Linguistics.

\bibitem[{Lee et~al.(2020)Lee, Lee, Jeong, Kim, and Hwang}]{lee2020generating}
Dong~Bok Lee, Seanie Lee, Woo~Tae Jeong, Donghwan Kim, and Sung~Ju Hwang. 2020.
\newblock \href {https://doi.org/10.18653/v1/2020.acl-main.20} {Generating
  diverse and consistent {QA} pairs from contexts with information-maximizing
  hierarchical conditional {VAE}s}.
\newblock In \emph{Proceedings of the 58th Annual Meeting of the Association
  for Computational Linguistics}, pages 208--224, Online. Association for
  Computational Linguistics.

\bibitem[{Li et~al.(2019)Li, Gao, Bing, King, and Lyu}]{li2019improving}
Jingjing Li, Yifan Gao, Lidong Bing, Irwin King, and Michael~R. Lyu. 2019.
\newblock \href {https://doi.org/10.18653/v1/D19-1317} {Improving question
  generation with to the point context}.
\newblock In \emph{Proceedings of the 2019 Conference on Empirical Methods in
  Natural Language Processing and the 9th International Joint Conference on
  Natural Language Processing (EMNLP-IJCNLP)}, pages 3216--3226, Hong Kong,
  China. Association for Computational Linguistics.

\bibitem[{Li et~al.(2016{\natexlab{a}})Li, Galley, Brockett, Gao, and
  Dolan}]{li2016diversity}
Jiwei Li, Michel Galley, Chris Brockett, Jianfeng Gao, and Bill Dolan.
  2016{\natexlab{a}}.
\newblock \href {https://doi.org/10.18653/v1/N16-1014} {A diversity-promoting
  objective function for neural conversation models}.
\newblock In \emph{Proceedings of the 2016 Conference of the North {A}merican
  Chapter of the Association for Computational Linguistics: Human Language
  Technologies}, pages 110--119, San Diego, California. Association for
  Computational Linguistics.

\bibitem[{Li et~al.(2016{\natexlab{b}})Li, Galley, Brockett, Spithourakis, Gao,
  and Dolan}]{li2016persona}
Jiwei Li, Michel Galley, Chris Brockett, Georgios Spithourakis, Jianfeng Gao,
  and Bill Dolan. 2016{\natexlab{b}}.
\newblock \href {https://doi.org/10.18653/v1/P16-1094} {A persona-based neural
  conversation model}.
\newblock In \emph{Proceedings of the 54th Annual Meeting of the Association
  for Computational Linguistics (Volume 1: Long Papers)}, pages 994--1003,
  Berlin, Germany. Association for Computational Linguistics.

\bibitem[{Li et~al.(2017)Li, Miller, Chopra, Ranzato, and
  Weston}]{li2016learning}
Jiwei Li, Alexander~H Miller, Sumit Chopra, Marc'Aurelio Ranzato, and Jason
  Weston. 2017.
\newblock Learning through dialogue interactions by asking questions.
\newblock \emph{ICLR}.

\bibitem[{Liu et~al.(2016)Liu, Lowe, Serban, Noseworthy, Charlin, and
  Pineau}]{liu2016not}
Chia-Wei Liu, Ryan Lowe, Iulian Serban, Mike Noseworthy, Laurent Charlin, and
  Joelle Pineau. 2016.
\newblock \href {https://doi.org/10.18653/v1/D16-1230} {How {NOT} to evaluate
  your dialogue system: An empirical study of unsupervised evaluation metrics
  for dialogue response generation}.
\newblock In \emph{Proceedings of the 2016 Conference on Empirical Methods in
  Natural Language Processing}, pages 2122--2132, Austin, Texas. Association
  for Computational Linguistics.

\bibitem[{Majumder et~al.(2020)Majumder, Jhamtani, Berg-Kirkpatrick, and
  McAuley}]{majumder2020}
Bodhisattwa~Prasad Majumder, Harsh Jhamtani, Taylor Berg-Kirkpatrick, and
  Julian McAuley. 2020.
\newblock \href {https://doi.org/10.18653/v1/2020.emnlp-main.739} {Like hiking?
  you probably enjoy nature: Persona-grounded dialog with commonsense
  expansions}.
\newblock In \emph{Proceedings of the 2020 Conference on Empirical Methods in
  Natural Language Processing (EMNLP)}, pages 9194--9206, Online. Association
  for Computational Linguistics.

\bibitem[{Nema et~al.(2019)Nema, Mohankumar, Khapra, Srinivasan, and
  Ravindran}]{nema2019let}
Preksha Nema, Akash~Kumar Mohankumar, Mitesh~M. Khapra, Balaji~Vasan
  Srinivasan, and Balaraman Ravindran. 2019.
\newblock \href {https://doi.org/10.18653/v1/D19-1326} {Let{'}s ask again:
  Refine network for automatic question generation}.
\newblock In \emph{Proceedings of the 2019 Conference on Empirical Methods in
  Natural Language Processing and the 9th International Joint Conference on
  Natural Language Processing (EMNLP-IJCNLP)}, pages 3314--3323, Hong Kong,
  China. Association for Computational Linguistics.

\bibitem[{Pang et~al.(2016)Pang, Lan, Guo, Xu, Wan, and Cheng}]{pang2016text}
Liang Pang, Yanyan Lan, Jiafeng Guo, Jun Xu, Shengxian Wan, and Xueqi Cheng.
  2016.
\newblock Text matching as image recognition.
\newblock In \emph{Thirtieth AAAI Conference on Artificial Intelligence}.

\bibitem[{Papineni et~al.(2002)Papineni, Roukos, Ward, and
  Zhu}]{papineni2002bleu}
Kishore Papineni, Salim Roukos, Todd Ward, and Wei-Jing Zhu. 2002.
\newblock \href {https://doi.org/10.3115/1073083.1073135} {{B}leu: a method for
  automatic evaluation of machine translation}.
\newblock In \emph{Proceedings of the 40th Annual Meeting of the Association
  for Computational Linguistics}, pages 311--318, Philadelphia, Pennsylvania,
  USA. Association for Computational Linguistics.

\bibitem[{Park et~al.(2018)Park, Cho, and Kim}]{park2018hierarchical}
Yookoon Park, Jaemin Cho, and Gunhee Kim. 2018.
\newblock \href {https://doi.org/10.18653/v1/N18-1162} {A hierarchical latent
  structure for variational conversation modeling}.
\newblock In \emph{Proceedings of the 2018 Conference of the North {A}merican
  Chapter of the Association for Computational Linguistics: Human Language
  Technologies, Volume 1 (Long Papers)}, pages 1792--1801, New Orleans,
  Louisiana. Association for Computational Linguistics.

\bibitem[{Serban et~al.(2016)Serban, Sordoni, Bengio, Courville, and
  Pineau}]{serban2016building}
Iulian~V Serban, Alessandro Sordoni, Yoshua Bengio, Aaron Courville, and Joelle
  Pineau. 2016.
\newblock Building end-to-end dialogue systems using generative hierarchical
  neural network models.
\newblock In \emph{Proceedings of the Thirtieth AAAI Conference on Artificial
  Intelligence}, pages 3776--3783.

\bibitem[{Serban et~al.(2017)Serban, Sordoni, Lowe, Charlin, Pineau, Courville,
  and Bengio}]{serban2017hierarchical}
Iulian~Vlad Serban, Alessandro Sordoni, Ryan Lowe, Laurent Charlin, Joelle
  Pineau, Aaron Courville, and Yoshua Bengio. 2017.
\newblock A hierarchical latent variable encoder-decoder model for generating
  dialogues.
\newblock In \emph{Proceedings of the Thirty-First AAAI Conference on
  Artificial Intelligence}, pages 3295--3301.

\bibitem[{Shang et~al.(2015)Shang, Lu, and Li}]{shang2015neural}
Lifeng Shang, Zhengdong Lu, and Hang Li. 2015.
\newblock \href {https://doi.org/10.3115/v1/P15-1152} {Neural responding
  machine for short-text conversation}.
\newblock In \emph{Proceedings of the 53rd Annual Meeting of the Association
  for Computational Linguistics and the 7th International Joint Conference on
  Natural Language Processing (Volume 1: Long Papers)}, pages 1577--1586,
  Beijing, China. Association for Computational Linguistics.

\bibitem[{Shen and Feng(2020)}]{shen2020cdl}
Lei Shen and Yang Feng. 2020.
\newblock \href {https://doi.org/10.18653/v1/2020.acl-main.52} {{CDL}:
  Curriculum dual learning for emotion-controllable response generation}.
\newblock In \emph{Proceedings of the 58th Annual Meeting of the Association
  for Computational Linguistics}, pages 556--566, Online. Association for
  Computational Linguistics.

\bibitem[{Shen et~al.(2019)Shen, Feng, and Zhan}]{shen2019modeling}
Lei Shen, Yang Feng, and Haolan Zhan. 2019.
\newblock \href {https://doi.org/10.18653/v1/P19-1549} {Modeling semantic
  relationship in multi-turn conversations with hierarchical latent variables}.
\newblock In \emph{Proceedings of the 57th Annual Meeting of the Association
  for Computational Linguistics}, pages 5497--5502, Florence, Italy.
  Association for Computational Linguistics.

\bibitem[{Shen et~al.(2021)Shen, Zhan, Shen, and Feng}]{shen2021learning}
Lei Shen, Haolan Zhan, Xin Shen, and Yang Feng. 2021.
\newblock Learning to select context in a hierarchical and global perspective
  for open-domain dialogue generation.
\newblock In \emph{ICASSP 2021-2021 IEEE International Conference on Acoustics,
  Speech and Signal Processing (ICASSP)}, pages 7438--7442. IEEE.

\bibitem[{Shen et~al.(2018)Shen, Su, Li, and Klakow}]{shen2018nexus}
Xiaoyu Shen, Hui Su, Wenjie Li, and Dietrich Klakow. 2018.
\newblock \href {https://doi.org/10.18653/v1/D18-1463} {{NEXUS} network:
  Connecting the preceding and the following in dialogue generation}.
\newblock In \emph{Proceedings of the 2018 Conference on Empirical Methods in
  Natural Language Processing}, pages 4316--4327, Brussels, Belgium.
  Association for Computational Linguistics.

\bibitem[{Tao et~al.(2018)Tao, Mou, Zhao, and Yan}]{tao2018ruber}
Chongyang Tao, Lili Mou, Dongyan Zhao, and Rui Yan. 2018.
\newblock Ruber: An unsupervised method for automatic evaluation of open-domain
  dialog systems.
\newblock In \emph{Thirty-Second AAAI Conference on Artificial Intelligence}.

\bibitem[{Tian et~al.(2017)Tian, Yan, Mou, Song, Feng, and Zhao}]{tian2017make}
Zhiliang Tian, Rui Yan, Lili Mou, Yiping Song, Yansong Feng, and Dongyan Zhao.
  2017.
\newblock \href {https://doi.org/10.18653/v1/P17-2036} {How to make context
  more useful? an empirical study on context-aware neural conversational
  models}.
\newblock In \emph{Proceedings of the 55th Annual Meeting of the Association
  for Computational Linguistics (Volume 2: Short Papers)}, pages 231--236,
  Vancouver, Canada. Association for Computational Linguistics.

\bibitem[{Vanderwende(2008)}]{vanderwende2008importance}
Lucy Vanderwende. 2008.
\newblock The importance of being important: Question generation.
\newblock In \emph{Proceedings of the 1st Workshop on the Question Generation
  Shared Task Evaluation Challenge, Arlington, VA}.

\bibitem[{Wang et~al.(2019)Wang, Feng, Wang, and Zhang}]{wang2019answer}
Weichao Wang, Shi Feng, Daling Wang, and Yifei Zhang. 2019.
\newblock \href {https://doi.org/10.18653/v1/D19-1511} {Answer-guided and
  semantic coherent question generation in open-domain conversation}.
\newblock In \emph{Proceedings of the 2019 Conference on Empirical Methods in
  Natural Language Processing and the 9th International Joint Conference on
  Natural Language Processing (EMNLP-IJCNLP)}, pages 5066--5076, Hong Kong,
  China. Association for Computational Linguistics.

\bibitem[{Wang et~al.(2018)Wang, Liu, Huang, and Nie}]{wang2018learning}
Yansen Wang, Chenyi Liu, Minlie Huang, and Liqiang Nie. 2018.
\newblock \href {https://doi.org/10.18653/v1/P18-1204} {Learning to ask
  questions in open-domain conversational systems with typed decoders}.
\newblock In \emph{Proceedings of the 56th Annual Meeting of the Association
  for Computational Linguistics (Volume 1: Long Papers)}, pages 2193--2203,
  Melbourne, Australia. Association for Computational Linguistics.

\bibitem[{Xing et~al.(2018)Xing, Wu, Wu, Huang, and
  Zhou}]{xing2018hierarchical}
Chen Xing, Yu~Wu, Wei Wu, Yalou Huang, and Ming Zhou. 2018.
\newblock Hierarchical recurrent attention network for response generation.
\newblock In \emph{Proceedings of the Thirty-Second AAAI Conference on
  Artificial Intelligence}, pages 5610--5617.

\bibitem[{Yu et~al.(2016)Yu, Xu, Black, and Rudnicky}]{yu2016strategy}
Zhou Yu, Ziyu Xu, Alan~W Black, and Alexander Rudnicky. 2016.
\newblock Strategy and policy learning for non-task-oriented conversational
  systems.
\newblock In \emph{Proceedings of the 17th annual meeting of the special
  interest group on discourse and dialogue}, pages 404--412.

\bibitem[{Zhan et~al.(2021)Zhan, Zhang, Chen, Ding, Bao, and
  Lan}]{zhan-etal-2021-augmenting}
Haolan Zhan, Hainan Zhang, Hongshen Chen, Zhuoye Ding, Yongjun Bao, and Yanyan
  Lan. 2021.
\newblock \href {https://www.aclweb.org/anthology/2021.naacl-main.446}
  {Augmenting knowledge-grounded conversations with sequential knowledge
  transition}.
\newblock In \emph{Proceedings of the 2021 Conference of the North American
  Chapter of the Association for Computational Linguistics: Human Language
  Technologies}, pages 5621--5630, Online. Association for Computational
  Linguistics.

\bibitem[{Zhang et~al.(2018)Zhang, Cui, Wang, Zhu, Li, Zhou, and
  Liu}]{zhang2018context}
Weinan Zhang, Yiming Cui, Yifa Wang, Qingfu Zhu, Lingzhi Li, Lianqiang Zhou,
  and Ting Liu. 2018.
\newblock Context-sensitive generation of open-domain conversational responses.
\newblock In \emph{Proceedings of the 27th International Conference on
  Computational Linguistics}, pages 2437--2447.

\bibitem[{Zhao et~al.(2017)Zhao, Zhao, and Eskenazi}]{zhao2017learning}
Tiancheng Zhao, Ran Zhao, and Maxine Eskenazi. 2017.
\newblock \href {https://doi.org/10.18653/v1/P17-1061} {Learning
  discourse-level diversity for neural dialog models using conditional
  variational autoencoders}.
\newblock In \emph{Proceedings of the 55th Annual Meeting of the Association
  for Computational Linguistics (Volume 1: Long Papers)}, pages 654--664,
  Vancouver, Canada. Association for Computational Linguistics.

\bibitem[{Zhou et~al.(2018)Zhou, Huang, Zhang, Zhu, and
  Liu}]{zhou2018emotional}
Hao Zhou, Minlie Huang, Tianyang Zhang, Xiaoyan Zhu, and Bing Liu. 2018.
\newblock Emotional chatting machine: Emotional conversation generation with
  internal and external memory.
\newblock In \emph{Proceedings of the AAAI Conference on Artificial
  Intelligence}, volume~32.

\bibitem[{Zhou et~al.(2019)Zhou, Zhang, and Wu}]{zhou2019question}
Wenjie Zhou, Minghua Zhang, and Yunfang Wu. 2019.
\newblock \href {https://doi.org/10.18653/v1/D19-1622} {Question-type driven
  question generation}.
\newblock In \emph{Proceedings of the 2019 Conference on Empirical Methods in
  Natural Language Processing and the 9th International Joint Conference on
  Natural Language Processing (EMNLP-IJCNLP)}, pages 6032--6037, Hong Kong,
  China. Association for Computational Linguistics.

\end{thebibliography}

\end{document}